\documentclass[sigconf]{acmart}

\AtBeginDocument{%
	}

\setcopyright{acmcopyright}
\copyrightyear{2022}
\acmYear{2022}
\acmDOI{10.1145/3503161.3548292}

\acmConference[MM '22] {Proceedings of the 30th ACM International Conference on Multimedia }{October 10--14, 2022}{Lisboa, Portugal.}
\acmBooktitle{Proceedings of the 30th ACM International Conference on Multimedia (MM '22), Oct. 10--14, 2022, Lisboa, Portugal}
\acmPrice{15.00}
\acmISBN{978-1-4503-9203-7/22/10}

\usepackage{latexsym}
\usepackage[T1]{fontenc}
\usepackage[utf8]{inputenc}
\usepackage{microtype}
\usepackage{graphicx}
\usepackage{amsmath}
\usepackage{amssymb}
\usepackage{multirow}
\usepackage{subfig}
\usepackage{balance}
\begin{document}
	
	\title{CMAL: A Novel Cross-Modal Associative Learning Framework for Vision-Language Pre-Training}
	
	
 \author{Zhiyuan Ma}
\orcid{0000-0002-0055-6485}
\affiliation{
	\institution{School of Computer Science and Technology, \\Huazhong University of Science and Technology, China}
 \country{}}
\email{zhiyuanma@hust.edu.cn}

\author{Jianjun Li}
\orcid{0000-0002-5265-7624}
\authornote{Corresponding author. This work was supported by the National Natural Science Foundation of China under Grant No 61672252.}
\affiliation{
	\institution{School of Computer Science and Technology, \\Huazhong University of Science and Technology, China}
 \country{}}
\email{jianjunli@hust.edu.cn}

\author{Guohui Li}
\affiliation{
	\institution{School of Computer Science and Technology, \\Huazhong University of Science and Technology, China}
 \country{}}
\email{guohuili@hust.edu.cn}

\author{Kaiyan Huang}
\orcid{0000-0001-7142-4730}
\affiliation{
	\institution{School of Computer Science and Technology, \\Huazhong University of Science and Technology, China}
 \country{}}
\email{kaiyanhuang@hust.edu.cn}

	\renewcommand{\shortauthors}{Zhiyuan Ma, Jianjun Li, Guohui Li, \& Kaiyan Huang}
	\begin{abstract}
		With the ﬂourishing of social media platforms, vision-language pre-training  (VLP) recently has received great attention and many remarkable progresses have been achieved. The success of VLP largely beneﬁts from the information complementation and enhancement between different modalities. However, most of recent studies focus on cross-modal contrastive learning (CMCL) to promote image-text alignment by pulling embeddings of positive sample pairs together while pushing those of negative pairs apart, which ignores the natural asymmetry property between different modalities and requires large-scale image-text corpus to achieve arduous progress. To mitigate this predicament, we propose CMAL, a \underline{C}ross-\underline{M}odal \underline{A}ssociative \underline{L}earning framework with anchor points detection and cross-modal associative learning for VLP. Specifically, we first respectively embed visual objects and textual tokens into separate hypersphere spaces to learn intra-modal hidden features, and then design a cross-modal associative prompt layer to perform anchor point masking and swap feature filling for constructing a hybrid cross-modal associative prompt. Afterwards, we exploit a unified semantic encoder to learn their cross-modal interactive features for context adaptation. Finally, we design an associative mapping classification layer to learn potential associative mappings between modalities at anchor points, within which we develop a fresh self-supervised  associative mapping classification  task to boost CMAL's performance. 
		Experimental results verify the effectiveness of CMAL, showing that it achieves competitive performance against previous CMCL-based methods on four common downstream vision-and-language tasks, with significantly fewer corpus. Noteably, CMAL obtains new state-of-the-art results on SNLI-VE and REC (testA). 
	\end{abstract}

	
	
	
	\maketitle
	
	\section{Introduction}

\begin{figure}[t]
	\centering
	\includegraphics[width=\linewidth,height=1.1\linewidth]{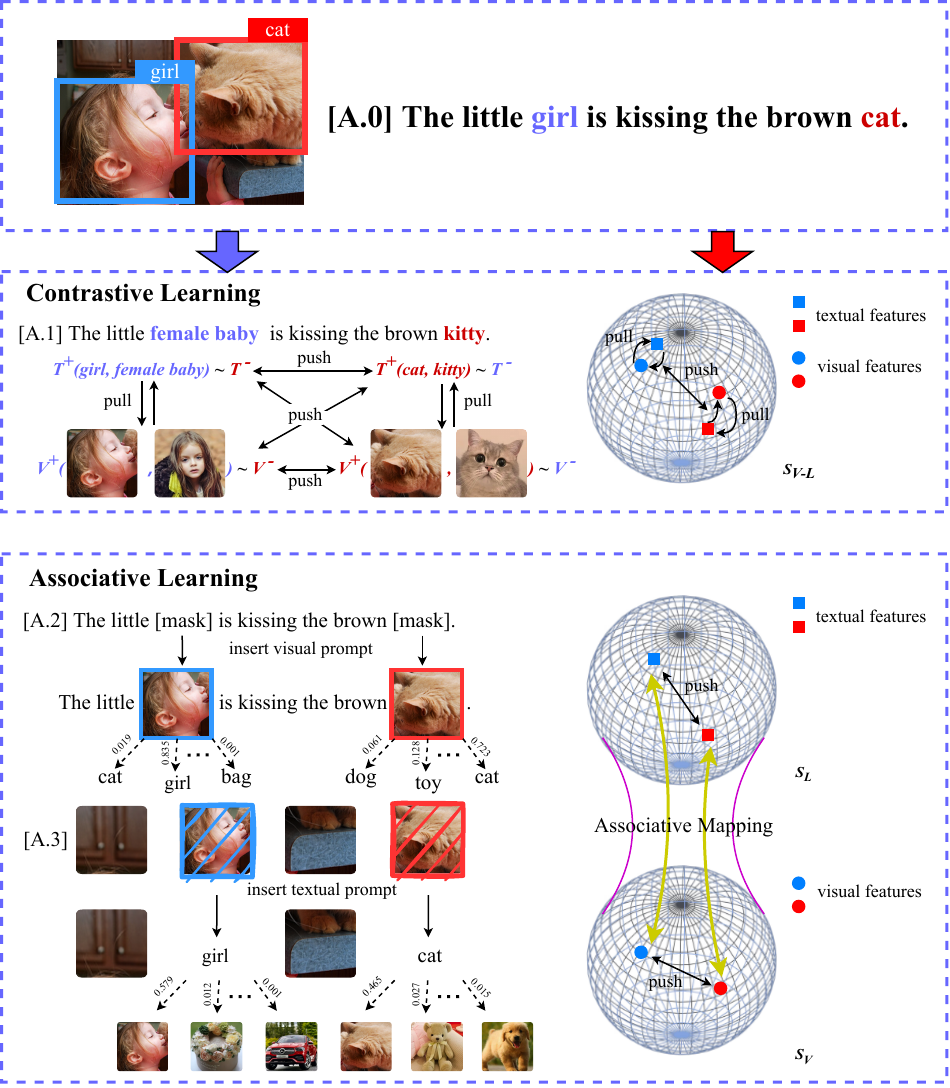}
	\vspace{-2em}
	\caption{Contrastive learning and associative learning for vision-language pretraining.}
	\label{fig: example}
	\vspace{-1.5em}
\end{figure}

Vision-Language Pre-training (VLP) aims to learn cross-modal semantic representations from large-scale image-text pairs to benefit various downstream vision-and-language (V+L) tasks, such as visual question answering~\cite{yu2017multi,yu2018beyond}, image text retrieval~\cite{lee2018stacked,plummer2015flickr30k}, and referring expression comprehension~\cite{kazemzadeh2014referitgame,yu2018rethinking}. Inspired by the success of Transformer and self-supervised pre-training, we have recently witnessed a boosting number of research works on VLP~\cite{chen2020uniter,li2020oscar,huang2020pixel,xu2021e2e,li2021unimo,li2021semvlp,luo2022vc,radford2021learning,li2022blip,li2019visualbert}. As a core study, learning fine-grained semantic alignments through contrastive learning plays a key role in VLP, which generally benefits from three crucial principles~\cite{wang2020understanding,chen2020simple}: (1) \emph{alignment} (closeness) of features from positive pairs, (2) \emph{uniformity} (normality) of the introduced distribution of the features from negative samples, and (3) \emph{global} scope to distinguish objects by introducing a large number of contrastive samples. Recently, many models based on cross-modal contrastive learning (CMCL) have been proposed~\cite{udandarao2020cobra,radford2021learning,li2021align,li2022mvp,li2021unimo,jia2021scaling,yang2022vision,li2020oscar,zhang2021vinvl}, which generally exploit an image-text contrastive (ITC) task to learn sentence-level semantic alignments~\cite{li2021align,radford2021learning,li2021unimo,jia2021scaling} or additionally exploit a word-region contrastive (WRC) task to learn token-level semantic alignments~\cite{yang2022vision,udandarao2020cobra,li2020oscar,zhang2021vinvl,li2022mvp} for better performance.

Though achieving remarkable progress, existing vision-language pre-training models still suffer from the following three limitations. (1) \emph{Asymmetry}. \textbf{Firstly}, images and texts are inherently asymmetric media, where image often contains richer and more implicit semantic information, while text usually contains more specific and explicitly focused description. Take the textual token \emph{``girl''}  and its corresponding visual object in Figure 1 as an  example, we believe that although they all refer to the same object \emph{girl}, the visual feature also contains descriptions of the little girl's appearance, such as curly hair and plump face, which suggests semantic representations of different modalities are inconsistent in feature dimensions. However, previous CMCL-based VLP models generally ignore the natural gaps between these two modalities, essentially treating them as the same modality to embed visual and textual features into a unified semantic space $S_{V\raisebox{0mm}{-}L}$.  Subsequently, cross-modal semantic alignments cannot be well captured and the first \emph{alignment} principle of contrast learning actually cannot be applied, which limits these models' ability to learn better cross-modal representations.  
(2) \emph{Non-normality}. \textbf{Secondly}, prior models generally use randomly chosen samples in the same batch to construct negative sample pairs (e.g., \emph{``girl''} and \emph{``cat''}).  Due to the lack of valuable negative sample pairs, especially those with similar attributes but different types, it is difficult to train the model to learn effective fine-grained representations that can distinguish subtle differences between different types of objects. 
One possible solution is manually constructing negative  pairs, but it also requires sufficient prior knowledge and a lot of time and effort, making it practically infeasible. Obviously, existing non-manually methods do not satisfy the second \emph{uniformity} principle of contrast learning, which limits their ability to represent different objects. 
(3) \emph{Locality}. \textbf{Thirdly}, limited by the size of the negative sample sets, prior CMCL-based methods in general can only perceive and represent local entities  
by pushing away the distances between negative sample pairs in the local scope, which violates the  \emph{global} principle of contrastive learning and makes it difficult to learn better global representations.

To address the aforementioned limitations, we propose a Cross-Modal Associative Learning (CMAL) framework for vision-language pre-training. Specifically, to address the first limitation, motivated by the asymmetry property and human brains' associative thinking, 
we employ a pipeline method to first detect the respective and common concerns (i.e., anchor points) in the image and text, and then stand on the shoulder of dual-stream Transformer-based models~\cite{DBLP:conf/emnlp/TanB19,DBLP:conf/nips/LuBPL19} to propose a hierarchical semantic encoder, which first projects the detected visual objects and textual words into separate semantic spaces to learn intra-modal features, and then exploits a unified semantic encoder to prompt inter-modal interactions, with the objective of learning better cross-modal representations. Based on the hierarchical encoder and detected anchor points, we further address the second limitation of negative sampling by designing a cross-modal associative prompt layer to perform anchor point masking and swap feature filling for obtaining more fine-grained representations of anchor points. Finally, to address the third limitation of locality, we devise a cross-modal associative classification layer to learn potential associative mappings by leveraging a fresh associative mapping classfication (AMC) loss to optimize the model, aiming at obtaining a better global representation for anchor point.  Experiments on four representative downstream V+L tasks demonstrate the effectiveness of CMAL, showing that it achieves competitive performance  against previous VLP models, but with significantly fewer pre-train corpus and lower computational cost. Particularly,  CMAL achieves new state-of-the-art results on two of the above four tasks,  SNLI-VE and REC (testA).   


	\section{Related Work}


\noindent\textbf{Unimodal Pre-training (UP).} The pre-training technique has been widely explored in both computer vision (CV) and natural language processing (NLP), due to its strong capability of generalization and efficient usage of large-scale data. In CV, a series of models based on deep convolutional networks, such as VGGNet~\cite{DBLP:journals/corr/SimonyanZ14a}, ResNet~\cite{he2016deep} and AlexNet~\cite{krizhevsky2017imagenet}, have been proposed and pretrained on ImageNet, and can well generalize to various downstream tasks~\cite{he2017mask,long2015fully,ren2015faster}, which effectively improves the capability of image recognition and object detection. With the widespread popularity of models based on the deep Transformer architecture, \emph{``pre-training + fine-tuning''} has been extended to NLP and many famous pretraining models (e.g., BERT~\cite{kenton2019bert}, GPT~\cite{radfordimproving} and XLNet~\cite{yang2019xlnet}) have been developed and shown promising results. Different from the aforementioned \emph{supervised pretraining} paradigm in CV tasks, the pretraining paradigm in NLP tasks is \emph{self-supervised} that aims to train a model to predict masked words based on their contexts without introducing human annotations. Recently, with the development of Vision Transformer (ViT)~\cite{dosovitskiy2020image}, such a \emph{self-supervised} paradigm has also been applied in CV area and  a series of ViT-based models~\cite{liu2021swin,wang2021pyramid,yuan2021tokens} have been proposed and applied in various CV tasks, showing great potential.\\

\begin{figure*}[t]
	\centering
	\includegraphics[width=0.85\linewidth]{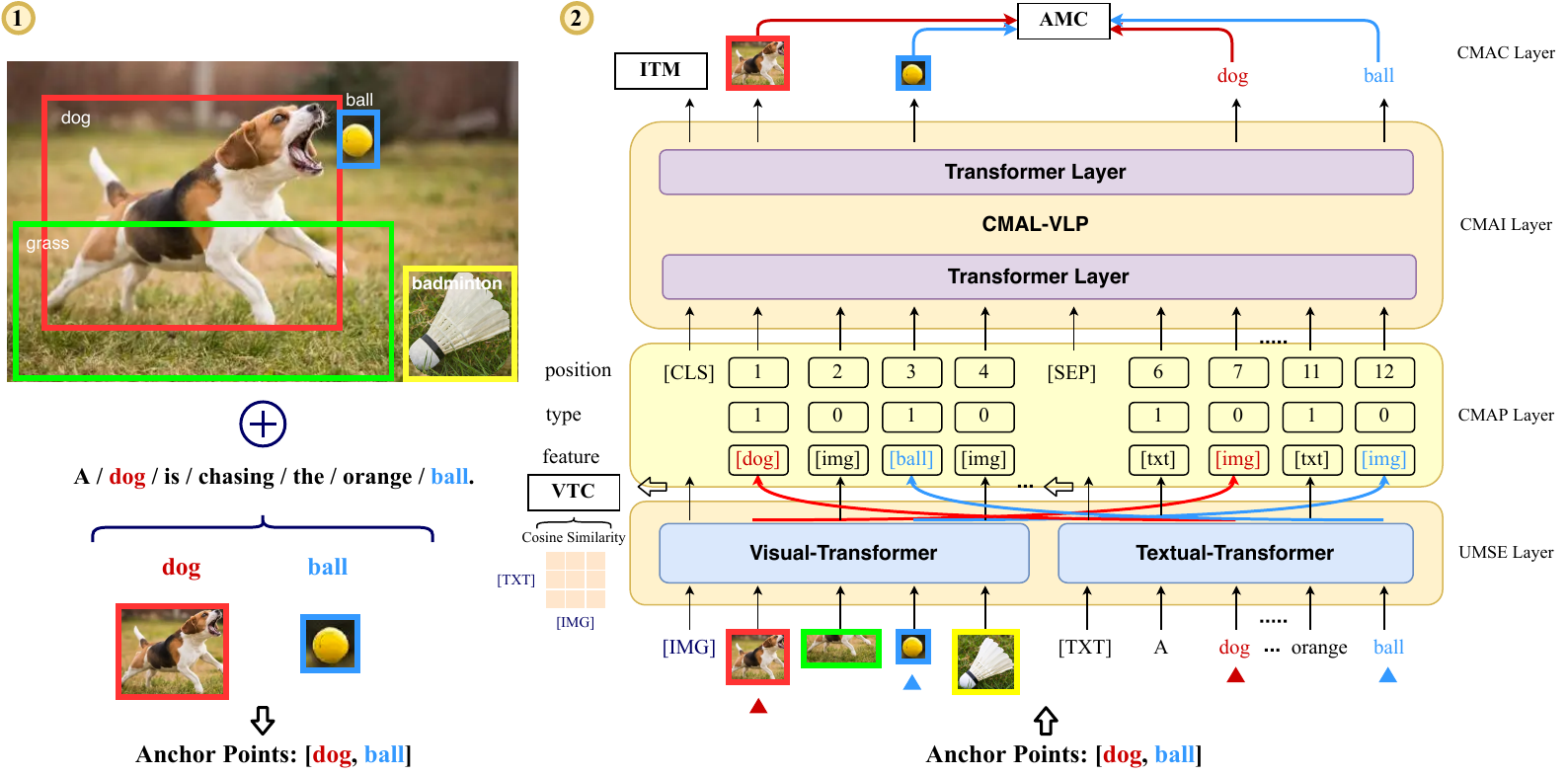}
	\vspace{-1em}
	\caption{The Proposed Framework.}
	\label{fig: model}
	\vspace{-0.8em}
\end{figure*}

\noindent\textbf{Vision-Language Pre-training (VLP).} Recent years has witnessed the remarkable success of unimodal pre-training in CV or NLP. With the flourishing of social media platforms, massive amounts of multimedia data are generated daily, which poses great demand for VLP. Different from the purely self-supervised paradigm in NLP tasks, VLP models are pretrained on large-scale paired image-text corpus. Based on the successful experience of the MLM strategy introduced by BERT~\cite{kenton2019bert} on purely text tasks, recent studies have also applied MLM to VLP models. Another MRM strategy similar to MLM is proposed to \emph{mask-then-predict} regions of images, and has been widely used for various V+L tasks~\cite{zhuge2021kaleido,zhou2021uc2,zhou2021learning,zhou2020unified,zhang2021vinvl,yu2021ernie,yang2022vision,li2021unsupervised,huang2021seeing,hong2021gilbert,cui2021rosita,DBLP:conf/iclr/SuZCLLWD20,dou2021empirical}. \textbf{ViLBERT}~\cite{DBLP:conf/nips/LuBPL19} and \textbf{LXMERT}~\cite{DBLP:conf/emnlp/TanB19} are two pioneering researches in this field, where the dual-stream architecture is adopted to encode visual features and textual features with two separate Transformers and then perform multimodal fusion via a unified Transformer. Different from them, recent works~\cite{DBLP:conf/aaai/LiDFGJ20,chen2020uniter,kim2021vilt,cui2021rosita} tend to use the single-stream architecture, where the multimodal features are directly fused using one Transformer, and show promising performance. 
\\

\noindent\textbf{Contrastive Learning for VLP (CL-VLP).} Though achieving remarkable progress, the performance of existing VLP models is still limited by cross-modal fine-grained alignments. Contrastive learning,  a recent popular representation learning technique, has been widely used in VLP to fill this gap. As a pioneering work, 
\textbf{COBRA}~\cite{udandarao2020cobra} presents a novel contrastive bi-modal representation algorithm that aims to train two modalities (i.e., image and text) in a joint fashion, supervised by the Contrastive Predictive Coding (CPC) and Noise Contrastive Estimation (NCE) loss. 
\textbf{OSCAR}~\cite{li2020oscar} is the first to introduce object tags into text-image sequence, it constructs a binary contrastive loss to learn tag-word alignments by randomly replacing object tags. Later, \textbf{VinVL}~\cite{zhang2021vinvl} extends the binary contrastive loss into $3\raisebox{0mm}{-}way$ contrastive loss to effectively transfer
 to VQA and text-image retrieval tasks. \textbf{CLIP}~\cite{radford2021learning} further proposes to use Visual Semantic Contrastive (VSC) task to train 400 million image-text pairs collected from the internet for VLP. Since the visual tokens and word tokens are still unaligned at a fine-grained level, \textbf{ALBEF}~\cite{li2021align} introduces a contrastive loss to align the image and text representations before fusing them through cross-modal attention. \textbf{UNIMO}~\cite{li2021unimo} further leverages cross-modal contrastive learning (CMCL) to align the textual and visual information into a unified semantic space. Likewise, \textbf{ALIGN}~\cite{jia2021scaling} also exploits such a contrastive loss to pull the embeddings of matched image-text pairs together while pushing those of non-matched image-text pairs apart. Very recently, \textbf{MVPTR}~\cite{li2022mvp} considers multi-level fine-grained semantic alignments into vision-language pre-training via a VSC loss and a Weakly-supervised Phrase Grounding (WPG) loss. Meanwhile, \textbf{TCL}~\cite{yang2022vision} takes advantage of localized and structural information and proposes triple contrastive learning for VLP by leveraging both cross-modal and intra-modal contrastive objective to provide complementary benefits in representation learning.  

Despite remarkable success, most of the existing VLP models ignore the asymmetry between modalities and rely heavily on contrastive learning to achieve image-text alignments for VLP. Different from them, our CMAL model  can learn cross-modal associative features by swapping the position of visual objects and textual tokens and mapping the swapped features into a global space to improve the performance of vision-language pre-training.  
	
	\section{Methodology}

As depicted in Figure~\ref{fig: model}, CMAL mainly contains two steps: 1) Anchor point detection (Sec.~\ref{subsec:CMAL-APD}) and 2) Vision-language pre-training (Sec.~\ref{subsec:CMAL-VLP}),  The former aims to find common concerns in image-text pairs (i.e., anchor points), while the latter aims to conduct visual-language pre-training based on the detected anchor points. All the pre-training objectives we used are described in Sec.~\ref{subsec:PTO}.

In CMAL, we design a fresh self-supervised pre-training associative mapping classification (AMC) task for cross-modal associative learning. This task is defined as predicting the most likely visual and textual token pairs $Y=\{y^v, y^t\}$, given their swapped features as cross-modal associative prompts. More specifically, given an image region sequence $X^v=\{x^v_1, x^v_2, \cdots, x^v_m\}$, a text tokenization sequence $X^t=\{x^t_1, x^t_2, \cdots, x^t_n\}$ and a detected anchor points set $\mathbb{A}$, the cross-modal prediction probability can be defined as,
\begin{equation}
	P(Y|M^v, M^t)=\sum_{(x^t_j, x^v_i)\in \mathbb{A}}^{|\mathbb{A}|} P\left(\{y^v_i, y^t_j\}\bigg| X^v_{\backslash i}\parallel x^t_j, X^t_{\backslash j}\parallel x^v_i\right)
\end{equation}
where $M^v$ and $M^t$ respectively represent the mixed sequence $X^v_{\backslash i}\parallel x^t_j$ and $X^t_{\backslash j}\parallel x^v_i$ after the features of anchor points are swapped. Here ``$\parallel$'' refers to the latter element being mixed into the previous sequence. More clearly, the two sequences after swapping are,
\begin{equation}
	M^v=\{x^v_1, \cdots, x^v_{i-1}, x^t_j, x^v_{i+1}, \cdots, x^v_m\}
\end{equation}
\begin{equation}
	M^t=\{x^t_1, \cdots, x^t_{j-1}, x^v_i, x^t_{j+1}, \cdots, x^t_n\}
\end{equation}
To obtain better intra-modal features, we do not directly exchange the above two sequences at the corresponding anchor points in the beginning, but first employ a UMSE layer (Sec.~\ref{subsubsec:UMSE})  to embed visual features and textual features into separate semantic spaces to learn their intra-modal representations. Then, we employ a CMAP layer (Sec.~\ref{subsubsec:CMAP}) to re-organize the intra-modal features into a hybrid sequence (similar to $M^v$ and $M^t$) as cross-modal associative prompts, by swapping their hidden state features from previous UMSE layer at anchor points. Considering that the self-attention mechanism in Transformer is order-less, we further introduce additional position embeddings and type embeddings  to respectively represent the position of current feature in the sequence and the modal category, so as to ensure the effectiveness of cross-modal associative learning. After that, the output of CMAP is fed into the CMAI layer (Sec.~\ref{subsubsec:CMAI}) to learn cross-modal associative interactions. 

Note unlike previous unified-modal VLP models~\cite{li2021unimo,chen2020uniter,DBLP:conf/aaai/LiDFGJ20,li2020oscar,zhou2020unified} based on image-text concatenation, our CMAI layer mainly serves the context adaptation of hybrid prompts for obtaining better context-aware cross-modal representations for anchor points. Finally, the anchor points' representations integrated with context information are output to the last CMAC layer  (Sec.~\ref{subsubsec:CMAC})  for cross-modal associative mapping classification.

\subsection{Anchor Point Detection}
\label{subsec:CMAL-APD}
The anchor point detection step is performed to detect separate and common concerns in the image-text pairs and generate aforementioned image region sequence $X^v=\{x^v_1, x^v_2, \cdots, x^v_m\}$, text tokenization sequence $X^t=\{x^t_1, x^t_2, \cdots, x^t_n\}$ and anchor point set $\mathbb{A}$, where $X^v$ and $X^t$ are used as inputs for subsequent VLP steps, while $\mathbb{A}$ is used to construct cross-modal associative prompts and perform fine-grained cross-modal associative learning in VLP.

Specifically, we first follow~\cite{chen2020uniter} to use Faster-RCNN\footnote{The Faster R-CNN we used was pre-trained on Visual Genome dataset~\cite{anderson2018bottom}.} to extract visual features (pooled RoI features) for each detected image region by region proposal network and serialize them into image region sequence $X^v$, then we obtain the object probability distribution of each region via the classifier of Faster-RCNN, and take the classification result with maximum probability as the object tag of the region. Here, the tag set corresponding to sequence $X^v$ is denoted as $\mathbb{T}_{X^v}$. Moreover, we also use a Bert Tokenizer to split the text sequence $X^t$ into words and exploit the split words to form a word set $\mathbb{W}_{X^t}$. Afterwards, we define a new anchor operator ``$\oplus$'', which is used to calculate the common concerns of region sequence $X^v$ and tokenization sequence $X^t$. Formally,
\begin{equation}
\label{formula:A}
	\mathbb{A}=X^v \oplus X^t=\mathbb{T}_{X^v} \cap \mathbb{W}_{X^t}
\end{equation}

Note that there are a lot of different expressions of synonymous objects in both region tags and split words, such as ``car'' and ``Car'', ``dog'' and ``dogs'', ``carpet'' and ``rug'', ``laptop'' and ``laptop computer'', ``CD'' and ``compact disk'', which in general can be classified into five categories, including case, singular and plural, synonyms, sub-words and abbreviations. Therefore, we  manually annotate all the objects in the pre-train dataset according to the above five categories to construct a thesaurus. Based on such a thesaurus, we standardize the above tag set $\mathbb{T}_{X^v}$ and the word set $\mathbb{W}_{X^t}$ by replacing variant words with the same form, to obtain a relatively high quality anchor point set $\mathbb{A}$ for subsequent vision-language pre-training.

\subsection{Vision-Language Pre-training}
\label{subsec:CMAL-VLP}

The vision-language pre-training step is designed to achieve cross-modal associative learning via the following four neural layers. 

\subsubsection{\textbf{Uni-Modal Semantic Encoding (UMSE)}}
\label{subsubsec:UMSE} 

To learn uni-modal semantic representations, we respectively adopt a single-layer Visual-Transformer and Textual-Transformer as encoders.

\paragraph{\textbf{Visual Semantic Encoder}}~In visual semantic encoder, we first encode the location features for each region of visual sequence $X^v$ via a 7-dimensional vector\footnote{The 7-dimensional location vector is consistent with UNITER~\cite{chen2020uniter}, i.e., $[x_1,y_1,x_2,y_2,w,h,w*h]$(normalized top/left/bottom/right coordinates, width, height, and area.)}, then feed the location vectors and original visual features into a fully-connected (FC) layer to obtain the final location-aware visual embeddings, denoted as $\textbf{V}$. Note here $\textbf{V}=(\textbf{v}_1,\textbf{v}_2,\cdots,\textbf{v}_m)$. Next, we concatenate a special token [IMG] before $\textbf{V}$ and combine them together as the input to the Visual-Transformer. Finally, we exploit the Visual-Transformer to produce image-modal contextual hidden state $\textbf{h}^v_i$, as well as the whole image semantic representation $\textbf{h}_{\text{[IMG]}}$,
\begin{equation}
	\textbf{h}^v_i=f_v(\textbf{v}_i), \quad   \textbf{h}_{\text{[IMG]}}=f_v(\text{[IMG]})
\end{equation}
where $f_v(\cdot)$ denotes the visual semantic encoder.

\paragraph{\textbf{Textual Semantic Encoder}}~In textual semantic encoder, similarly, we first sequentially encode the position features of each word in text sequence $X^t$, then feed the position vectors and original textual features into a fully-connected (FC) layer to obtain the final position-aware word embeddings, denoted as $\textbf{W}$. Note here $\textbf{W}=(\textbf{w}_1,\textbf{w}_2,\cdots,\textbf{w}_n)$. Next, we concatenate a special token [TXT] before $\textbf{W}$ and combine them together as the input to the Textual-Transformer. Finally, we exploit the Textual-Transformer to produce text-modal contextual hidden state $\textbf{h}^t_j$, as well as the whole sentence semantic representation $\textbf{h}_{\text{[TXT]}}$,
\begin{equation}
\textbf{h}^t_j=f_t(\textbf{t}_j), \quad   \textbf{h}_{\text{[TXT]}}=f_t(\text{[TXT]})
\end{equation}
where $f_t(\cdot)$ denotes the textual semantic encoder. 


\subsubsection{\textbf{Cross-Modal Associative Prompt (CMAP)}}
\label{subsubsec:CMAP}
The CMAP layer is the key to associative learning, which aims to construct a cross-modal prompt template by swapping visual and textual features at anchor points. Motivated by recent prompt learning technology~\cite{liu2021pre}, we first define a \emph{masking function} $g_{\text{mask}}(\cdot)$ through anchor point masking, and then process the input visual sequence $\textbf{h}^v$ and textual sequence $\textbf{h}^t$ into a mask prompt $\textbf{H}_{\text{prompt}}=g_{\text{mask}}(\textbf{h}^v, \textbf{h}^t)$. We also define a \emph{filling function} $g_{\text{fill}}(\cdot)$ through swap feature filling and process them into a hybrid swapped sequence $\textbf{H}_{\text{swap}}=g_{\text{fill}}(\textbf{H}_{\text{prompt}})$ based on the anchor point set $\mathbb{A}$.

\paragraph{\textbf{Anchor point masking}} In this stage, the CMAL model randomly scans  set $\mathbb{A}$ and masks a scanned anchor point's feature in both visual feature sequence $\textbf{h}^v$ and textual feature sequence $\textbf{h}^t$ with a certain probability\footnote{The mask probability of the anchor point is set to $100\%$ in CMAL.}. Note that only one randomly scanned anchor point is masked at each time and each anchor point is masked only once. Specifically, for $\textbf{h}^v=(\textbf{h}^v_1,\cdots,\textbf{h}^v_m)$ and $\textbf{h}^t=(\textbf{h}^t_1,\cdots,\textbf{h}^t_n)$, the masked prompt is as follows:
\begin{equation}
	\begin{aligned}
		\textbf{H}_{\text{prompt}}=\Big(&\text{[CLS]},\textbf{h}^v_1, \cdots, \textbf{h}^v_{i-1}, [\text{mask}]^v_i, \textbf{h}^v_{i+1}, \cdots, \textbf{h}^v_m,\Big.\\
		&\Big.\text{[SEP]},\textbf{h}^t_1, \cdots, \textbf{h}^t_{j-1}, [\text{mask}]^t_j, \textbf{h}^t_{j+1}, \cdots, \textbf{h}^t_n\Big)
	\end{aligned}
\end{equation}
where $[\text{mask}]^v_i$ and $[\text{mask}]^t_j$ respectively denote the $i$-th feature of visual sequence and the $j$-th feature of textual sequence corresponding to the currently scanned anchor point to be replaced by a special token [mask].

\paragraph{\textbf{Swap feature filling}} In this stage, the CMAL model fills the swapped cross-modal features in the corresponding ``[mask]'' position of template $\textbf{H}_{\text{prompt}}$ with a particular probability\footnote{By grid search on hyperparameters, the swap probability of anchor point is set to 60\% in CMAL.} to construct the hybrid prompt feature sequence $\textbf{H}_{\text{swap}}$ as follows:
\begin{equation}
	\begin{aligned}
		\textbf{H}_{\text{swap}}=\Big(&\text{[CLS]},\textbf{h}^v_1, \cdots, \textbf{h}^v_{i-1}, [\textbf{h}]^t_j, \textbf{h}^v_{i+1}, \cdots, \textbf{h}^v_m,\Big.\\
		&\Big.\text{[SEP]},\textbf{h}^t_1, \cdots, \textbf{h}^t_{j-1}, [\textbf{h}]^v_i, \textbf{h}^t_{j+1}, \cdots, \textbf{h}^t_n\Big)
	\end{aligned}
\end{equation}
where $[\textbf{h}]^t_j$ and $[\textbf{h}]^v_i$ respectively denote swapped features $\textbf{h}^t_j$ and $\textbf{h}^v_i$.

\paragraph{\textbf{Position, type, and feature integrating}} Since the self attention mechanism in Transformer is not sensitive to position, which may inactivate the hybrid prompt sequence $\textbf{H}_{\text{swap}}$, we integrate position embedding, modal type embedding and hybrid prompt feature by feeding them into a fully-connected (FC) layer, to project them into the same embedding space. The ﬁnal hybrid feature that integrates position and modal type is obtained by ﬁrst summing up the three FC outputs, and then passing them through an LN layer. Finally, the hybrid hidden feature sequence concatenated with a classification token [CLS] and a separator token [SEP] can be obtained as, 
\begin{equation}
	\begin{aligned}
		\textbf{H}_{\text{hybrid}}=\Big(&\text{[CLS]},\textbf{H}^v_{1:i-1}, \textbf{H}^t_j, \textbf{H}^v_{i+1:m},\Big.\text{[SEP]},\Big.\\ 
		&\Big.\textbf{H}^t_{m+2:m+j},\textbf{H}^v_i, \textbf{H}^t_{m+j+2:m+n+1}\Big)
	\end{aligned}
\end{equation}
where $\textbf{H}^v$ and $\textbf{H}^t$ respectively denote the visual and textual hidden state features after integrating position information and modal type features.  Note here $m$ and $n$ respectively represent the length of visual sequence and the length of textual sequence, while $i$ and $j$ respectively denote the $i$-th feature of visual sequence and the $j$-th feature of textual sequence swapped at the anchor point. Moreover, [CLS] stands for [CLS]$_{0}$ and [SEP] stands for [SEP]$_{m+1}$. The total length is $m+n+2$, i.e., the subscript starts at $0$ and ends at $m+n+1$.

\subsubsection{\textbf{Cross-Modal Associative Interaction (CMAI)}} 
\label{subsubsec:CMAI}
This layer is adopted to learn cross-modal associative interactions and make the above swapped anchor points' feature adapt to the changed context, so as to obtain a better context-aware cross-modal representation for subsequent associative mapping classification. Similar to BERT, we adopt 12 layers of Transformer as CMAL's backbone, and define the backbone by an \emph{interaction function} $f_{\text{inter}}(\cdot)$ as, 
\begin{equation}
	\textbf{H}^{(l)}_{\text{output}}=f^{(l)}_{\text{inter}}(\textbf{H}^{(l-1)}_{\text{output}})
\end{equation}
\begin{equation}
	\textbf{H}^{(0)}_{\text{output}}=\textbf{H}_{\text{hybrid}}
\end{equation}
where $l$ denotes layer index, and $\textbf{H}^{(12)}_{\text{output}}$ is the output of the last layer. Finally, the CMAI layer retrieves the swapped visual hidden feature $\textbf{H}^v_{anchor}$ and textual hidden feature $\textbf{H}^t_{anchor}$ at the anchor point from $\textbf{H}^{(12)}_{\text{output}}$ for the prediction of the next classification layer.

\subsubsection{\textbf{Cross-Modal Associative Classification (CMAC)}}
\label{subsubsec:CMAC}
The CMAC layer is designed to learn cross-modal associative mappings. Specifically, given hidden prompt feature $\textbf{H}^t_{anchor}$ and $\textbf{H}^v_{anchor}$, it is trained to  recover swapped visual region and textual word respectively by predicting the anchor point's category.

For textual recovering, CMAL first feeds hidden visual prompt feature $\textbf{H}^v_{anchor}$ into an MLP, and then uses a \emph{softmax} function to obtain the final word prediction probability $P(y^t_j| M^v,M^t)$. Formally,
\begin{equation}
	P(y^t_j|M^v,M^t)=\text{softmax}\left( \text{MLP}\left(\textbf{H}^v_{anchor}\right)\right)
\end{equation}
For visual recovering, we follow UNITER~\cite{chen2020uniter} to perform swapped region classification (SRC)\footnote{Similar to the masked region classification (MRC) in UNITER.}. Specifically, CMAL first feeds hidden textual prompt feature $\textbf{H}^t_{anchor}$ into an FC layer to predict the scores of $K$ object classes, and then further utilizes \emph{softmax}  to obtain a normalized probability distribution $g(y^v_i|M^v,M^t)$. Formally,
\begin{equation}
	g(y^v_i|M^v,M^t)=\text{softmax}\left( \text{FC}\left(\textbf{H}^t_{anchor}\right)\right)
\end{equation}

\subsection{Pre-Training Objective}
\label{subsec:PTO}
\textbf{Associative Mapping Classification (AMC).}~Our most critical pre-training objective is the AMC loss, which consists of two parts,
\begin{equation}
	\mathcal{L}_{\text{AMC}}=\mathcal{L}_{\text{TXT}}+\mathcal{L}_{\text{IMG}}
\end{equation}
Specifically, the textual recovering loss is defined as:
\begin{equation}
	\mathcal{L}_{\text{TXT}}(\theta)=-\mathbb{E}_{(\textbf{V},\textbf{W}) \sim D} \log P_{\theta}(y^t_j| M^v,M^t)
\end{equation}
where $\theta$ denotes all trainable parameters and each pair $(\textbf{V}, \textbf{W})$ is sampled from the whole training set $D$. The visual recovering loss is defined as:
\begin{equation}
	\mathcal{L}_{\text{IMG}}(\theta)=\mathbb{E}_{(\textbf{V},\textbf{W}) \sim D} \text{CE}\left( c(\textbf{H}^t_{anchor}),g_{\theta}(y^v_i|M^v,M^t)\right)
\end{equation}
where CE($\cdot$) represents for cross-entropy loss and $c(\textbf{H}^t_{anchor})$ denotes a category one-hot vector of the anchor point corresponding to $\textbf{H}^t_{anchor}$ in the object space of which Faster-RCNN detected.

\vspace{0.3em}
\noindent\textbf{Masked Language Modeling (MLM).}~To train CMAL's representation ability for general words (i.e., words not in the anchor set), we follow BERT to mask general words with a 15\% probability\footnote{Following BERT, we decompose this 15\% into 10\% random words, 10\% unchanged, and 80\% [MASK].}, and replace the masked word $\textbf{w}_m$ with special token [MASK]. The objective is to predict these masked general words based on the attention of their contextual words $\textbf{W}_{\backslash m}$ and all image regions $\textbf{V}$, by minimizing the negative log-likelihood:
\begin{equation}
	\mathcal{L}_{\text{MLM}}(\theta)=-\mathbb{E}_{(\textbf{V},\textbf{W}) \sim D} \log P_{\theta}(\textbf{w}_m|\textbf{W}_{\backslash m}, \textbf{V})
\end{equation} 

\vspace{0.3em}
\noindent\textbf{Masked Region Modeling (MRM).}~To train CMAL's representation ability for general regions (i.e., whose tags are not in the anchor set), we adopt a masked region classification (MRC) loss similar to the above visual recovering loss as the objective of MRM\footnote{The mask probability of MRM is the same as that of MLM.}. The objective is to predict these masked general regions $\textbf{v}_m$ based on the attention of their contextual regions $\textbf{V}_{\backslash m}$ and all textual words $\textbf{W}$, by minimizing the cross-entropy loss:
\begin{equation}
	\mathcal{L}_{\text{MRM}}(\theta)=-\mathbb{E}_{(\textbf{V},\textbf{W}) \sim D} \log P_{\theta}(\textbf{v}_m|\textbf{V}_{\backslash m}, \textbf{W})
\end{equation} 

\vspace{0.3em}
\noindent\textbf{Image-Text Matching (ITM).}~To train CMAL's image-text matching ability to better adapt to downstream tasks such as image-text retrieval, we first retrieve the last hidden output $\textbf{H}^{(12)}_{\text{[CLS]}}$ of [CLS] from $\textbf{H}^{(12)}_{\text{output}}$, and then feed it into an FC layer and a \emph{sigmoid} function to predict a probability between 0 and 1. We denote the output probability as $p_{\theta}(\textbf{W}, \textbf{V})$, the ITM loss is then defined as:
\begin{equation}
	\mathcal{L}_{\mathrm{ITM}}(\theta)=-\mathbb{E}_{(\textbf{V}, \textbf{W}) \sim D}[y \log p_{\theta}(\textbf{W}, \textbf{V}) + (1-y)
	\log \left(1-p_{\theta}(\textbf{W}, \textbf{V})\right)]
\end{equation}
where $y \in \{0,1\}$ is a truth binary label. 
 
 \vspace{0.3em}
\noindent\textbf{Vision-Text Contrast (VTC).}~Following previous work~\cite{radford2021learning,li2022mvp}, we build a contrastive learning-based version of our CMAL model by adding a VTC task. The VTC task aims to align image-text pairs of the same batch at a coarse-grained level, which is defined as: given a batch of $N$ image-text pairs, where $N$ image sequences are paired with $N$ text sequences respectively to obtain a total of $N$ positive sample pairs and $N^2-N$ negative sample pairs. The model is optimized by maximizing the cosine similarity of positive sample pairs and minimizing the cosine similarity of negative sample pairs. Specifically, the similarity $S_{i,j}$ between the image and text is computed by,
\begin{equation}
	S_{i,j}=\cos (\textbf{h}^i_{\text{[IMG]}},\textbf{h}^j_{\text{[TXT]}})
\end{equation}
while the image-to-text and text-to-image loss are computed by,
\begin{equation}
	\mathcal{L}_{\text{v2t}}(\textbf{h}^i_{\text{[IMG]}},\textbf{h}_{\text{[TXT]}})=-\sum^N_{k=1}\log\text{softmax}(S_{i,k}/\tau)
\end{equation}
\begin{equation}
	\mathcal{L}_{\text{t2v}}(\textbf{h}^j_{\text{[TXT]}},\textbf{h}_{\text{[IMG]}})=-\sum^N_{k=1}\log\text{softmax}(S_{j,k}/\tau)
\end{equation}
where $\tau$ is a temperature hyperparameter. Finally, the whole VTC loss is computed as,
\begin{equation}
	\small
	\mathcal{L}_{\mathrm{VTC}}(\theta)=\frac{1}{2N}\left(\sum^N_{i=1}\mathcal{L}_{\text{v2t}}(\textbf{h}^i_{\text{[IMG]}},\textbf{h}_{\text{[TXT]}})+\sum^N_{j=1}\mathcal{L}_{\text{t2v}}(\textbf{h}^j_{\text{[TXT]}},\textbf{h}_{\text{[IMG]}})\right)
\end{equation}

In sum, based on the above five pre-training tasks, our model is trained to minimize the total objective as:
\begin{equation}
	\mathcal{L}_{\text{Total}}=\gamma_{1}\mathcal{L}_{\text{AMC}}+\gamma_{2}\mathcal{L}_{\text{MLM}}+\gamma_{3}\mathcal{L}_{\text{MRM}}+\gamma_{4}\mathcal{L}_{\text{ITM}}+\gamma_{5}\mathcal{L}_{\text{VTC}}
\end{equation}
where $\gamma_{1-5}$ are hyperparameters that refer to the proportion of data supervised by the corresponding task in each pre-training epoch.

	\section{Experiments}
\setlength{\tabcolsep}{3pt}
\begin{table}[]
	\begin{tabular}{c|cc|cc|c}
		\hline \hline
		\multirow{2}{*}{\textbf{Split}} & \multicolumn{2}{c|}{\textbf{COCO captions}~\cite{chen2015microsoft} } & \multicolumn{2}{c|}{\textbf{VG captions}~\cite{krishna2017visual}} & \multirow{2}{*}{\textbf{Total}} \\ \cline{2-5}
		&    \# images                &  \# texts         &   \# images    &  \# texts   &        \\ \hline \hline
		train & 79K & 277K & 101K & 5.06M & 180K/5.337M \\
		valid & 5K  & 25K  & 2.1K  & 106K  & 7.1K/131K   \\ \hline \hline
	\end{tabular}
	\vspace{-1pt}
	\caption{Statistics on the datasets used for pre-training.}
	\label{tab: statistics}
	\vspace{-2.5em}
\end{table}

\setlength{\tabcolsep}{3pt}
\begin{table*}[]
	\small
	\begin{tabular}{l|c|cc|cc|cc|ccc}
		\hline \hline
		\multirow{2}{*}{\textbf{Method}} &
		\multirow{2}{*}{\textbf{Pre-train Corpus}  $(\#images / \#texts)$} &
		\multicolumn{2}{c|}{\textbf{VQA}} &
		\multicolumn{2}{c|}{\textbf{NLVR$^2$}} &
		\multicolumn{2}{c|}{\textbf{SNLI-VE}} &
		\multicolumn{3}{c}{\textbf{REC}} \\ \cline{3-11} 
		&                       & \textbf{dev}   & \textbf{test}  & \textbf{dev}   & \textbf{test}  & \textbf{dev}   & \textbf{test}  & \textbf{dev}   & \textbf{testA} & \textbf{testB} \\ \hline \hline
		\textbf{ViLBERT}~\cite{DBLP:conf/nips/LuBPL19}     &     CC (3.3M/3.3M)               & 70.55 & 70.92 & -   & -     &     -  &      - & 72.34 & 78.52 & 62.61 \\
		\textbf{VLBERT}~\cite{DBLP:conf/iclr/SuZCLLWD20}      &       CC (3.3M/3.3M) +BooksCorpus (74M)+English Wikipedia          & 70.50 & 70.83 & -      & -      &   -    &    -   & 71.60 & 77.72 & 60.99 \\
		\textbf{UNITER}~\cite{chen2020uniter}      &   COCO+VG+CC+SBU (4.197M/9.583M)                   & 72.27 & 72.46 & 75.85 & 75.80 & 78.59 & 78.28 & 75.31 & 81.30 & 65.58 \\
		\textbf{ViLT}~\cite{kim2021vilt}        &     COCO+VG+SBU+GCC (4.098M/9.854M)                  & 71.26 &     -  & 75.70 & 76.13 &    -   &  -     &    -   &   -    &   -    \\
		\textbf{VILLA}~\cite{gan2020large}       &    COCO+VG+CC+SBU (4.197M/9.583M)                     & 73.59 & 73.67 & \textbf{78.39} & \textbf{79.30} & \textbf{79.47} & \textbf{79.03} & 76.05 & 81.65 & 65.70 \\
		\textbf{ROSITA}~\cite{cui2021rosita}      & \multicolumn{1}{c|}{COCO+VG+CC+SBU (4.197M/9.583M)  } & \textbf{73.91} & \textbf{73.97} &    -   &    -   &     -  &    -   & \textbf{76.06} & \textbf{82.01} & \textbf{67.40}$^*$ \\ \hline
		\textbf{OSCAR}~\cite{li2020oscar}       &   COCO+CC+ SBU+GQA (4.602M/6.5M)                  & 73.16 & 73.44 & 78.07 & 78.36 &    -   &   -    &     -  &      - &   -    \\
		\textbf{UNIMO}~\cite{li2021unimo} &
		\multicolumn{1}{c|}{COCO+VG+CC+SBU (4.197M/9.583M) } &
		\multicolumn{1}{c}{73.79} &
		\multicolumn{1}{c|}{74.02} &
		\multicolumn{1}{c}{-} &
		\multicolumn{1}{c|}{-} &
		\multicolumn{1}{c}{80.00} &
		\multicolumn{1}{c|}{79.10} &
		\multicolumn{1}{c}{-} &
		\multicolumn{1}{c}{-} &
		\multicolumn{1}{c}{-} \\
		\textbf{ALBEF}~\cite{li2021align} &
		\multicolumn{1}{c|}{COCO+VG+CC12M+SBU (4M/5.146M) } &
		\multicolumn{1}{c}{74.54} &
		\multicolumn{1}{c|}{74.70} &
		\multicolumn{1}{c}{80.24} &
		\multicolumn{1}{c|}{80.50} &
		\multicolumn{1}{c}{80.14} &
		\multicolumn{1}{c|}{\textbf{80.30}} &
		\multicolumn{1}{c}{-} &
		\multicolumn{1}{c}{-} &
		\multicolumn{1}{c}{-} \\
		\textbf{VinVL}~\cite{zhang2021vinvl}       & \multicolumn{1}{c|}{COCO+CC+SBU+VQAs+OpenImages (6.035M/8.851M)} & 75.95 & 76.12 & \textbf{82.05}$^*$ & \textbf{83.08}$^*$ & -     & -     & 74.04 & 80.50 & 65.96 \\
		\textbf{TCL}~\cite{yang2022vision}         &      COCO+VG+SBU+CC+CC12M (14.962M/16.085M)                 & 74.90 & 74.92 & 80.54 & 81.33 & \textbf{80.51} & 80.29 & -     & -     & -     \\
		\textbf{MVPTR}~\cite{li2022mvp}       & \multicolumn{1}{c|}{COCO+CC+SBU+VQAs+OpenImages (6.035M/8.851M)} & \textbf{76.16}$^*$ & \textbf{76.36}$^*$ &    -   &  -     & 80.30 & 80.17 & \textbf{74.80} & \textbf{80.88} & \textbf{67.11} \\ \hline 
		
		\textbf{VisualBERT}~\cite{li2019visualbert}  &         COCO (106K/0.533M)             & 70.80 & 71.00 & 67.40 & 67.00 &  -     &    -   &     -  &  -     &   -    \\
		\textbf{LXMERT}~\cite{DBLP:conf/emnlp/TanB19}      & COCO+VG+VQA+GQA+VGQ (180K/9.18M)       & 72.42 & 72.54 & 74.90 & 74.50 &    -   &    -   &     -  &  -     &   -    \\
		\textbf{E2E-VLP}~\cite{xu2021e2e} &  
		\multicolumn{1}{c|}{COCO+VG (180K/6.01M)} &
		\multicolumn{1}{c}{73.25} &
		\multicolumn{1}{c|}{73.67} &
		\multicolumn{1}{c}{77.25} &
		\multicolumn{1}{c|}{77.96} &
		\multicolumn{1}{c}{-} &
		\multicolumn{1}{c|}{-} &
		\multicolumn{1}{c}{-} &
		\multicolumn{1}{c}{-} &
		\multicolumn{1}{c}{-} \\ \hline \hline
		\textbf{CMAL (Ours)}        &       COCO+VG (180K/5.337M)               & \textbf{73.41} & \textbf{73.84} & \textbf{78.30} & \textbf{80.02} & \textbf{81.30}$^*$ & \textbf{80.83}$^*$ & \textbf{76.92}$^*$ & \textbf{82.53}$^*$ & \textbf{66.72} \\ \hline \hline
	\end{tabular}
	\vspace{-1pt}
	\caption{Main results on four downstream tasks. Bold indicates the winner of each group and $*$ indicates the global winner. Note all our results are statistically significant with $p < 0.05$ under t-test.}
	\label{table: results}
	\vspace{-15pt}
\end{table*}

\subsection{Pre-training Setup}
\textbf{Datasets.} Following~\cite{xu2021e2e}, we construct our pre-training dataset consisting of 5.337M train and 131k validation image-text pairs from two public datasets, namely the COCO Captions~\cite{chen2015microsoft} and the Visual Genome (VG) Captions~\cite{krishna2017visual}. We only collect train and valid splits in each dataset to avoid seeing any test data in pre-training stage. Note the initial COCO dataset processed by UNITER~\cite{chen2020uniter} contains 106K images and 533K sentences, among which 27K images and 256K sentences are duplicated with the VG dataset. We remove the duplicated data and obtain 79K images and 277K texts to train. The statistics of the pre-training datasets are summarized in Table~\ref{tab: statistics}. 

\vspace{0.5em}
\noindent\textbf{Implementation Details.} 
 CMAL  adopts $12$-layer Transformers as backbone. The size of the tokenized word set $\mathbb{W}_{X^t}$ is $28997$, the size of the object tag set $\mathbb{T}_{X^v}$ is $1600$, and the size of the computed anchor point set $\mathbb{A}$ is $1269$, which means $1269$ meta-mappings between objects and words are contained. 
The initial learning rate is set as $5e^{-5}$, and the weight decay is set as $0.01$. The propotion hyperparameters $\gamma_{1-5}$ are set to $[0.4, 0.2, 0.2, 0.1, 0.1]$ respectively. Note the hyperparameters are all tuned with grid-search over the validation set. We use the AdamW~\cite{loshchilov2018decoupled} optimizer to optimize the model and all experiments are performed on $2$ NVIDIA A100 GPUs with PyTorch framework. Our code and pre-trained models are released at: \url{https://github.com/AnonymousPony/cmal}.

\vspace{0.5em}
\noindent\textbf{Baselines.}~For a more holistic and objective comparison, we compare CMAL versus the latest 15 SOTA models, and classify them into three groups: 1) pre-training on large-scale images based on non-contrastive learning, including {ViLBERT}~\cite{DBLP:conf/nips/LuBPL19}, {VLBERT}~\cite{DBLP:conf/iclr/SuZCLLWD20}, {UNITER}~\cite{chen2020uniter}, {ViLT}~\cite{kim2021vilt}, {VILLA}~\cite{gan2020large} and {ROSITA}~\cite{cui2021rosita}; 2) pre-training on large-scale images based on contrastive learning, including {OSCAR}~\cite{li2020oscar}, {UNIMO}~\cite{li2021unimo}, {ALBEF}~\cite{li2021align}, {VinVL}~\cite{zhang2021vinvl}, {TCL}~\cite{yang2022vision} and {MVPTR}~\cite{li2022mvp}; 3) pre-training on small-scale images based on non-contrastive learning, including {VisualBERT}~\cite{li2019visualbert}, {LXMERT}~\cite{DBLP:conf/emnlp/TanB19}, {E2E-VLP}~\cite{xu2021e2e}. All baselines use the \emph{base} model for a fair comparison. Note our model CMAL can be classified into group 3.

\subsection{Downstream Tasks}
After obtaining the pre-trained CMAL model, we finetune it on  four representative downstream V+L tasks, as stated below. 

\vspace{0.5em}
\noindent\textbf{Visual Question Answering (VQA)}~\cite{yu2017multi} aims to answer natural language questions about an image. We evaluate our model on VQAv2 dataset~\cite{goyal2017making}, which is manually built on the images from the MSCOCO dataset~\cite{lin2014microsoft}. The dataset is split into training (83K images and 444K questions), validation (41K images and 214K questions), and test (81K images and 448K questions) sets. We consider this task as a generation problem by following the same setting in~\cite{li2021align}. Specifically, an answer decoder is fine-tuned to generate the answer from the 3129 candidates.

\vspace{0.5em}
\noindent\textbf{Visual Reasoning (NLVR$^2$)}~\cite{suhr2019corpus} aims to determine whether a natural language caption is true about a pair of photographs. We adopt the widely used NLVR$^2$ dataset to evaluate our model, which contains 107292 image-caption samples with 86K/7K data for training/development. Since this task takes a text and two images as input, we follow the UNITER~\cite{chen2020uniter} to process them into two pairs and perform joint reasoning to obtain prediction results. 

\vspace{0.5em}
\noindent\textbf{Visual Entailment (SNLI-VE)}~\cite{xie2019visual} aims to judge whether a given image semantically entails a given text and give out an \emph{entailment}, \emph{contradiction}, or \emph{neutral} judgment. The dataset is split into training (30K images and 529K sentences), validation (1K images and 18K sentences), and test (1K images and 18K questions) sets. Compared with VQA and NLVR$^2$, this task requires fine-grained reasoning to reach a reliable conclusion.

\vspace{0.5em}
\noindent\textbf{Referring Expression Comprehension (REC)}~\cite{kazemzadeh2014referitgame} aims to localize an image region referred to by a natural language query. We evaluate our model on RefCOCO+~\cite{kazemzadeh2014referitgame} to test the performance in referring expression comprehension. The dataset is collected from COCO~\cite{plummer2015flickr30k} and is split into train set with 120k queries, valid set with 11k queries, testA set with 6k queries about people, and testB set with 6k queries about objects. The task is formulated as a binary classification problem and optimized through cross-entropy loss.

\setlength{\tabcolsep}{2.5pt}
\begin{table}[]
	\small
	\begin{tabular}{l|l|c|c|c|c|c}
		\hline \hline
		\multicolumn{1}{c|}{\multirow{2}{*}{\textbf{\#}}} & \multicolumn{1}{c|}{\multirow{2}{*}{\textbf{Pre-training Tasks}}} & \multirow{2}{*}{\textbf{Sum}} & \textbf{VQA} & \textbf{NLVR$^2$} & \textbf{VE} & \textbf{REC} \\ \cline{4-7} 
		\multicolumn{1}{c|}{} & \multicolumn{1}{c|}{}      &        & \textbf{dev} & \textbf{dev}  & \textbf{dev}  & \textbf{dev} \\ \hline \hline
		1                     & None (directly fine-tuned) & 220.54 & 67.02    & 51.12 & 53.64 & 48.76 \\ \hline
		2                     & ITM                        & 226.41 & 67.63    & 52.37 & 52.25 & 54.16 \\ \hline
		3                     & ITM+MLM                    & 283.22 & 70.12    & 72.18 & 72.09 & 68.83 \\ \hline
		4                     & ITM+MLM+MRM                & 287.26 & 70.16    & 72.12 & 73.21 & 71.77 \\ \hline
		5                     & ITM+MLM+MRM+VTC            & 292.73 & 71.53    & 74.60 & 74.21 & 72.39 \\ \hline
		6                     & ITM+MLM+MRM+VTC+AMC        & \textbf{299.13} & \textbf{71.96}    & \textbf{75.62} & \textbf{77.35} & 
		\textbf{74.20} \\ \hline \hline
	\end{tabular}
	\vspace{-1pt}
	\caption{Ablation study on pre-training tasks. Note all our results are statistically significant with $p < 0.05$ under t-test.}
	\label{table: ablation}
	\vspace{-2em}
\end{table}

\subsection{Overall Performance}
We compare CMAL with existing SOTA baselines on the aforementioned four downstream tasks. As shown in Table~\ref{table: results}, CMAL achieves competitive performance against existing VLP models. Specifically, CMAL achieves significant improvement on all the four downstream tasks compared to a series of classic VLP models, such as ViLBERT, VLBERT and ViLT in group 1, as well as all the SOTA models in group 3, demonstrating the effectiveness of the proposed cross-modal associative learning paradigm. 
Note in order to support contrastive learning for performance improvement, almost all the CMCL-based VLP models in group 2 need to be pre-trained on millions of images. For example, MVPTR is pre-trained on $6.035$ million images, while TCL is pre-trained  even on $14.962$ million images. This, with no doubt, takes a lot of computational cost. In sharp contrast, CMAL can obtain comparable performance with only 180K images on VQA and NLVR$^2$, and even achieves new SOTA results on SNLI-VE and REC (testA), which  illustrates the practicality of our model.
As a landmark model in VLP, the pre-training tasks of UNITER, including  MLM, MRM, WRA and ITM, have been widely used by subsequent models. Compared with UNITER, CMAL achieves an improvement of 1.90\%, 5.57\%, 3.26\%, 1.51\% and 1.74\% in the 5 test sets respectively, but only uses 4\% images and 56\% sentences, which further verifies its effectiveness.

\subsection{Ablation Studies}
In this part, we perform ablation experiments to evaluate the effectiveness of different pre-training settings over the four downstream tasks. Besides standard metrics for each task, we also use Sum (sum of all the scores across the four tasks) as a global metric. We focus on six crucial pre-training settings, as shown in Table~\ref{table: ablation}. $\#1$ means there is no pre-training task, and we directly fine-tune the model in the downstream tasks; $\#2$ means we only use the most basic ITM task to pre-train our model; $\#3$ means an additional MLM task is used to observe the performances on the downstream tasks; $\#4$ means an additional MRM task is added to observe the performances of the model; $\#5$ means an additional VTC task is adopted to build a contrastive learning-based version; $\#6$ means the AMC task is included to build a complete CMAL model. Note that all the pre-training tasks are assigned the same weight in the ablation experiments to make a fair comparison. From Table~\ref{table: ablation}, we can observe that each addition of a pre-training task leads to an improvement in the model performance, which proves the effectiveness of all the pre-training tasks employed by CMAL. In particular, when the AMC task is included,  CMAL improves by 0.60\%, 1.37\%, 4.23\% and 2.50\% on the four downstream tasks respectively, compared with the pre-training set $\#5$, which further verifies the effectiveness of our cross-modal associative learning. Moreover, it can also be observed that CMAL obtains larger improvements on SNLI-VE and REC tasks, which is consistent with the results in our overall evaluation.

\begin{figure}[t]
	\centering
	\subfloat[\scriptsize{Associative mapping space $S_V\oplus S_L$.}]{
		\begin{minipage}[t]{0.22\textwidth}
			\flushleft
			\includegraphics[height=0.7\textwidth]{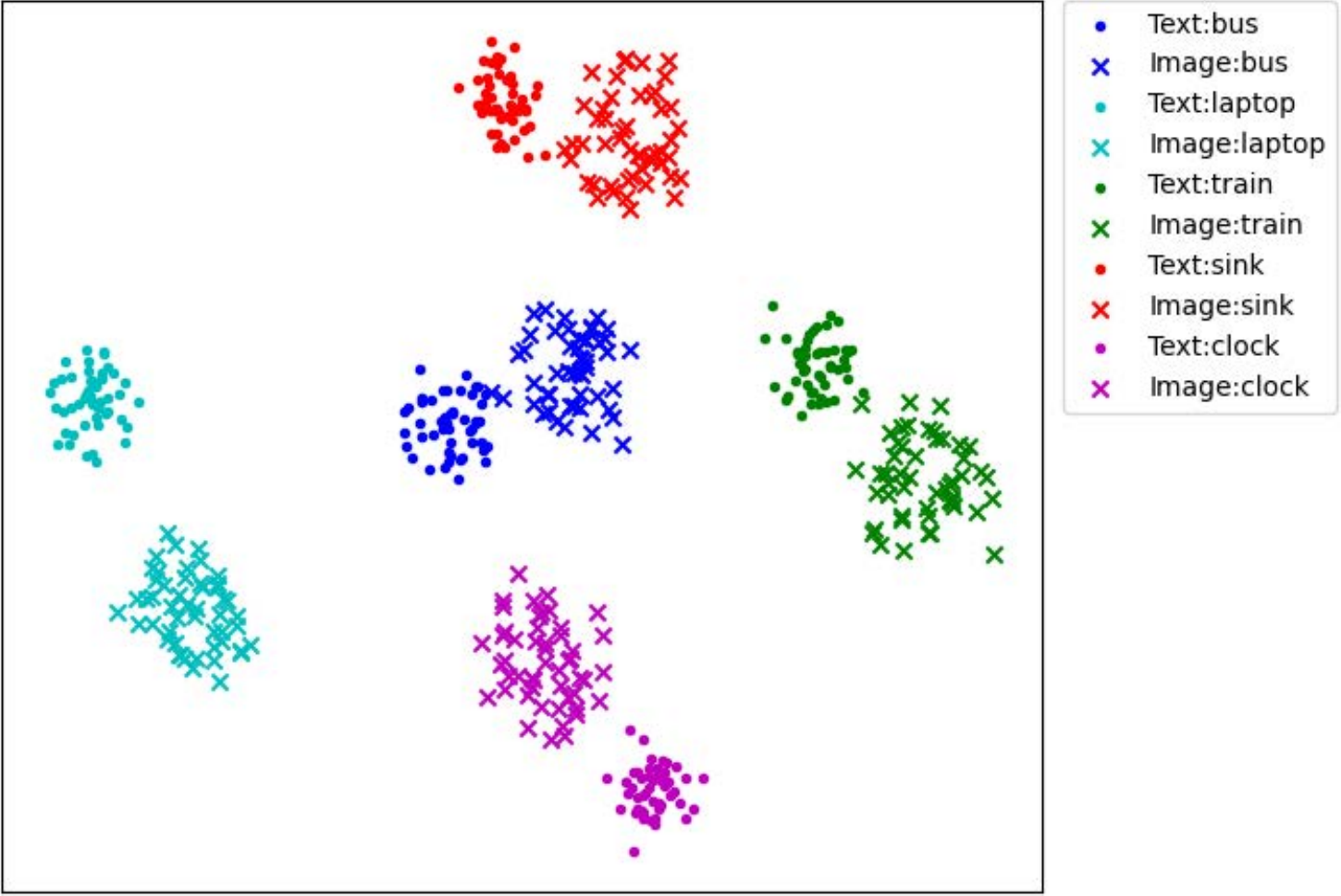}
			\vspace{-0.5em}
			\label{fig:associative mapping space}
		\end{minipage}
	}\hspace{0.2cm}
	\subfloat[\scriptsize{Contrastive learning space $S_{V\raisebox{0mm}{-}L}$.}]{
		\begin{minipage}[t]{0.22\textwidth}
			\flushright
			\includegraphics[height=0.7\textwidth]{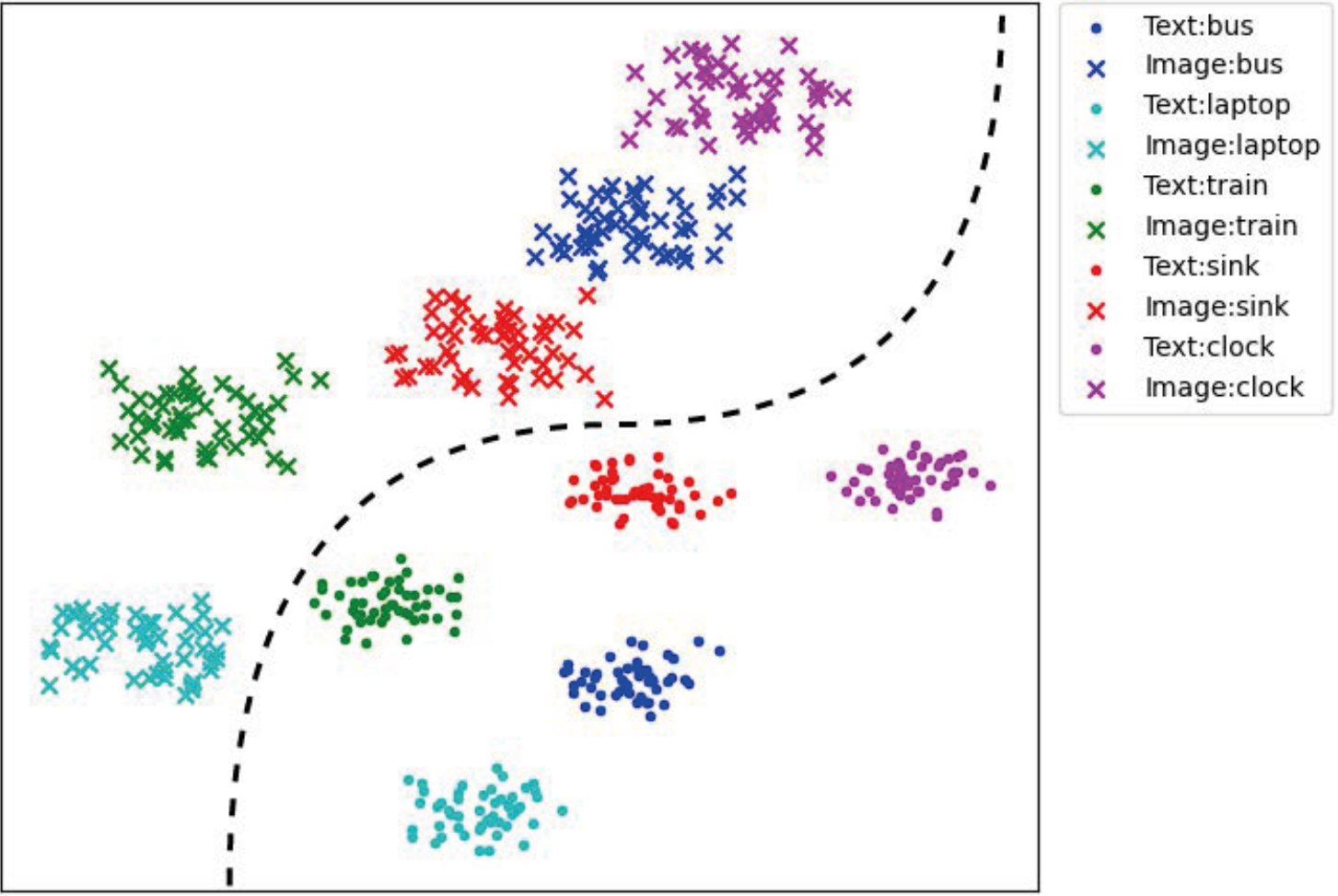}
				\vspace{-0.5em}
			\label{fig:contrastive learning space}
		\end{minipage}
	}\\
	\vspace{-5pt}
	\caption{2D visualization of the hidden features using $t$-SNE. The same anchor points share the same color.}
	\label{fig: cmap-visualization}
	\vspace{-2em}
\end{figure}

\begin{figure}[t]
	\centering
	\subfloat[\scriptsize{Associative learning.}]{
		\begin{minipage}[t]{0.24\textwidth}
			\flushleft
			\includegraphics[height=0.7\textwidth]{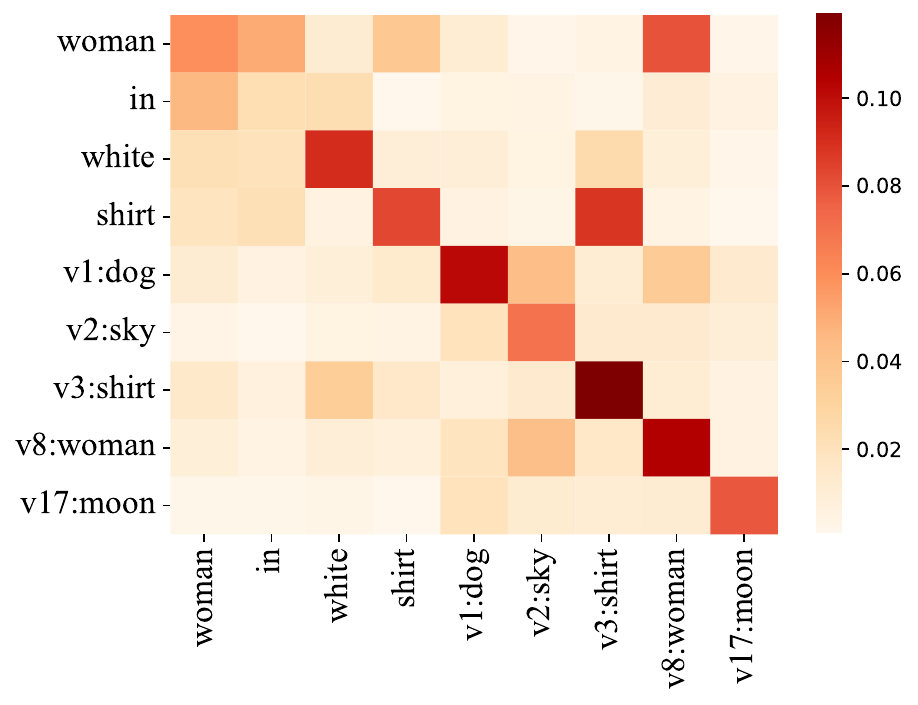}
		\end{minipage}
	}\hspace{-0.8cm}
	\subfloat[\scriptsize{Contrastive learning.}]{
		\begin{minipage}[t]{0.24\textwidth}
			\flushright
			\includegraphics[height=0.7\textwidth]{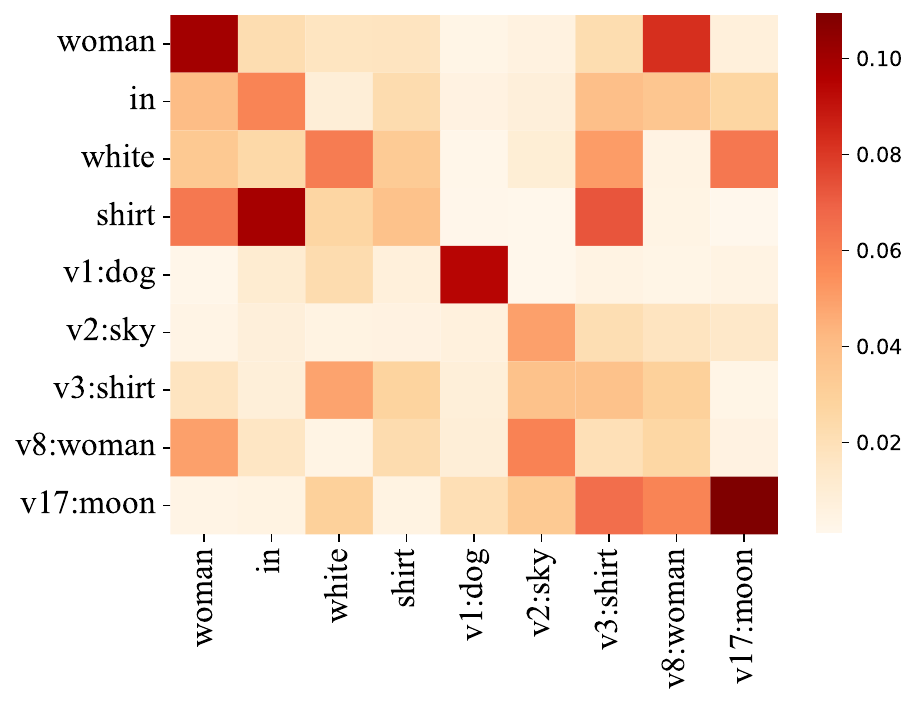}
		\end{minipage}
	}\\
	\vspace{-7pt}
	\caption{Attention visualization of associative learning versus contrastive learning.}
	\label{fig: cmai-visualization}
	\vspace{-1.3em}
\end{figure}

\subsection{Further Analysis}
\textbf{Cross-modal associative prompt.}~To more clearly illustrate the gap between the two different modalities and show how our CMAP layer works, we respectively visualize 5 anchor points' features from our CMAL and from the hidden layer of Transformer in a purely contrastive learning-based method, as shown in Figure~\ref{fig: cmap-visualization}. From~Figure~\ref{fig:contrastive learning space}, we can observe an obvious semantic gap between the image-modal and text-modal, due to their natural asymmetry property. From Figure~\ref{fig:associative mapping space}, we can see that CMAL can better align anchor points' visual features and textual features, which demonstrates the effectiveness of our cross-modal associative learning.

\vspace{0.5em}
\noindent\textbf{Cross-modal associative interaction.}~To better illustrate the advantage of our model in fine-grained learning and understand what the cross-modal associative interaction layer has learned, we respectively visualize the attention weights from CMAI layer and from the hidden layer of Transformer in a  contrastive learning-based method, as shown in Figure~\ref{fig: cmai-visualization}. It can be observed that our associative learning-based model is more adept at mining crucial semantic information (e.g., \emph{``woman''} and \emph{``shirt''}) and cross-modal semantic alignments between the two associative modal-features, and can effectively resist contextual noises due to being supervised by AMC task, while the contrastive learning-based method, which is limited by scattered attention weights from context (e.g., \emph{``white''} and \emph{``in''}), is hard to learn fine-grained semantic representations. \\

\begin{figure}[t]
	\centering
	\includegraphics[width=\linewidth,height=0.5\linewidth]{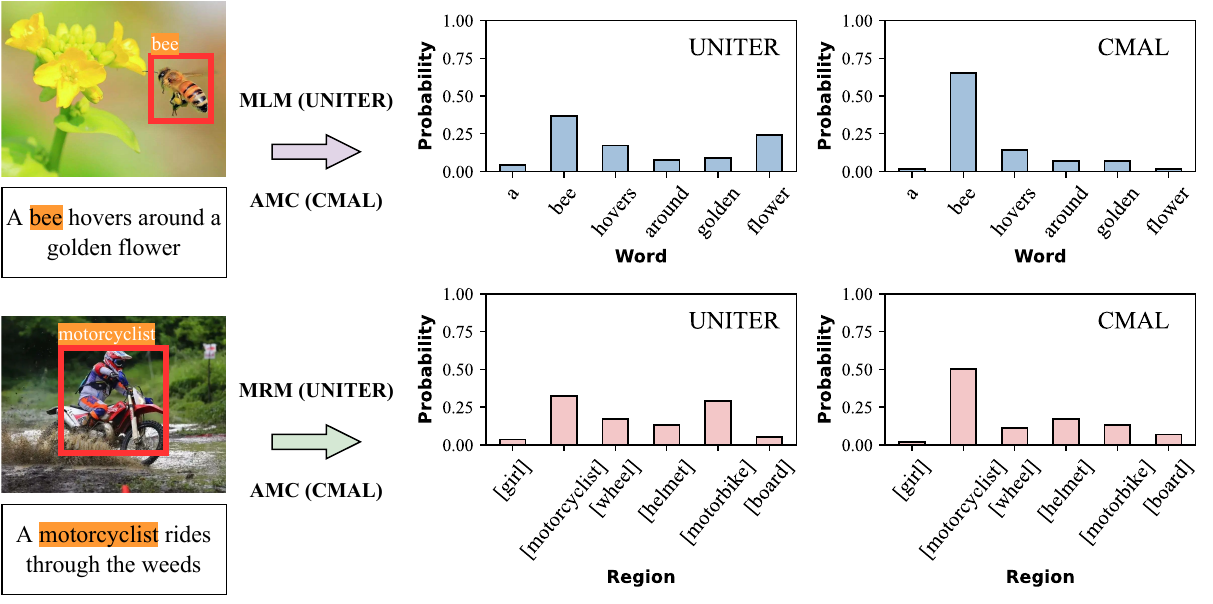}
	\vspace{-2em}
	\caption{Cross-modal associative classification.}
	\label{fig: cmac-visualization}
	\vspace{-1.5em}
\end{figure}

\noindent\textbf{Cross-modal associative classification.}~To further verify whether our model can effectively distinguish contextual words in text task as well as semantically relevant objects in image task and correctly classify them, we visualize the probability distribution from our CMAC layer and the classification layer of UNITER. From the visualized distributions in Figure~\ref{fig: cmac-visualization}, we can see that, compared with UNITER, CMAL can more correctly classify the anchor point's word (i.e., \emph{``bee''}) and region (i.e., \emph{``motorcyclist''}) after giving the cross-modal associative prompts. Specifically, when the word \emph{``bee''} is masked, by giving the swapped region features \emph{``[bee]''}, CMAL can accurately recover the word \emph{``bee''}, while UNITER may be disturbed by the surrounding words such as \emph{``flower''}. Moreover, this phenomenon is more obvious in image task. Since \emph{``motorcyclist''} and \emph{``motorbike''} are semantically related but actually different objects, it can be observed that they achieve very close classification probability in UNITER, while there is a large gap between them in CMAL. This verifies the stability of CMAL and demonstrates its superiority in obtaining global representations for anchor points.

	\section{Conclusion}

In this paper,  we propose CMAL, a cross-modal associative learning framework with anchor points detection and cross-modal associative learning for vision-language pre-training. Specifically, we first embed visual and textual features into separate semantic spaces to learn their intra-modal features, and then employ a unified semantic encoder to learn their inter-modal features. Moreover, a fresh self-supervised pre-training task, AMC, is proposed and used to improve the performance of the CMAL model. 
 Experiments on four well-known downstream V+L tasks demonstrate the effectiveness of CMAL, showing that it achieves competitive performance on four representative downstream tasks with fewer pre-train corpus and lower computational cost. Moreover, CMAL aslo obtains new state-of-the-art results on SNLI-VE and REC (testA) tasks, verifying its effectiveness.

	\vfill\eject
	\bibliographystyle{ACM-Reference-Format}
	\balance
	\bibliography{mm2022}

\end{document}